\documentclass[review]{elsarticle}

\usepackage{amsmath,amssymb,amsfonts}
\usepackage{algorithmic}
\usepackage{graphicx}
\usepackage{textcomp}
\usepackage{multirow}
\usepackage{epstopdf}
\usepackage{tabularx}
\usepackage{caption}
\usepackage[labelsep=period]{caption}
\usepackage{textcomp}
\usepackage{verbatim}
\usepackage{xcolor}
\usepackage{hyperref}
\journal{Neurocomputing}

\begin{document}

\begin{frontmatter}

\title{Edge and Identity Preserving Network\\ for Face Super-Resolution}
\tnotetext[mytitlenote]{This work was conducted when Jonghyun Kim visited Xidian University as a co-operating researcher.}
\tnotetext[mytitlenote]{Manuscript received Aug. 18, 2020; revised Feb. 17, 2021; Accepted Mar. 13, 2021.}

\author[1,2]{Jonghyun Kim}
\ead{jhkim.ben@gmail.com}
\address[1]{Department of Electronic, Electrical and Computer Engineering, Sungkyunkwan University, Suwon 16419, South Korea}
\author[1]{Gen Li}
\ead{ligen@skku.edu}
\author[3]{Inyong Yun}
\ead{iyyun@skku.edu}
\address[3]{Big Data \& AI Lab, Hana Institute of Technology, Hana TI, Seoul 06133, South Korea}
\author[2]{Cheolkon Jung\corref{corresponding author}}
\ead{zhengzk@xidian.edu.cn}
\cortext[corresponding author]{Corresponding author.}
\address[2]{School of Electronic Engineering, Xidian University, Xi'an 710071, China\vspace{-0.8cm}}
\author[1]{Joongkyu Kim\corref{corresponding author}}
\ead{jkkim@skku.edu}

\begin{abstract}
Face super-resolution (SR) has become an indispensable function in security solutions such as video surveillance and identification system, but the distortion in facial components is a great challenge in it. Most state-of-the-art methods have utilized facial priors with deep neural networks. These methods require extra labels, longer training time, and larger computation memory. In this paper, we propose a novel Edge and Identity Preserving Network for Face SR Network, named as EIPNet, to minimize the distortion by utilizing a lightweight edge block and identity information. We present an edge block to extract perceptual edge information, and concatenate it to the original feature maps in multiple scales. This structure progressively provides edge information in reconstruction to aggregate local and global structural information. Moreover, we define an identity loss function to preserve identification of SR images. The identity loss function compares feature distributions between SR images and their ground truth to recover identities in SR images. In addition, we provide a luminance-chrominance error (LCE) to separately infer brightness and color information in SR images. The LCE method not only reduces the dependency of color information by dividing brightness and color components but also enables our network to reflect differences between SR images and their ground truth in two color spaces of RGB and YUV. The proposed method facilitates the proposed SR network to elaborately restore facial components and generate high quality $8\times$ scaled SR images with a lightweight network structure. Furthermore, our network is able to reconstruct an $128\times128$ SR image with 215 fps on a GTX 1080Ti GPU. Extensive experiments demonstrate that our network qualitatively and quantitatively outperforms state-of-the-art methods on two challenging datasets: CelebA and VGGFace2.
\end{abstract}


\end{frontmatter}

%\linenumbers

\section{Introduction}

Face super-resolution (SR) is a challenging problem in computer vision, and various methods \cite{liu2018iterative,jiang2019atmfn,jiang2018context,FSRNet,exemplar,progressive,heatmaps,tiny, jiang2020dual, jiang2016srlsp, zeng2019face} have been proposed in recent years. A main goal of the face SR is to recognize human's identities, but it is difficult to restore high-frequency details in facial regions. For example, in an upscaling process from low-resolution (LR) face images to high resolution face images, face SR suffers from distortion of facial components such as eyes, nose and mouth. Therefore, the elaborate restoration of the facial components is necessary for real-world applications, i.e., security, video surveillance system, face recognition and identification.

Previously, face SR methods \cite{Edgeinform, SREdgeNet, seanet, jiang2018deep} utilized edge information to mitigate endemic distortion problems in an SR task. Such tendencies prove that the utilization of edges is an intuitive solution to improve SR images. Specifically, these methods can be divided into two steps which consist of edge extraction and SR conversion. Firstly, an edge network extracts edge information from low-resolution edge maps and coarse images. Secondly, the extracted edge information is conveyed to an SR network to compensate the high-frequency components, which leads to an increase in performance of the SR methods. However, they employ a sub-network, i.e., an encoder-decoder network structure, to generate edge maps and only consider the edge maps in a single scale. 

\begin{figure}[t]
\begin{center}
\begin{tabular}{cc}
\includegraphics[width=3.6cm]{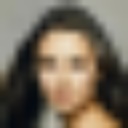}&
\includegraphics[width=3.6cm]{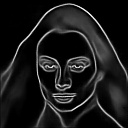}\\
Input(Bilinear)&Edge map
\vspace{0.1cm}\\
\includegraphics[width=3.6cm]{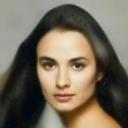}&
\includegraphics[width=3.6cm]{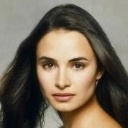}\\
Ours&Target
\end{tabular}
\caption{Edge map and super-resolution result by the proposed method with scale factor 8.}
\vspace{-0.3cm}
\label{figure1}
\end{center}
\end{figure}

\begin{figure}[t]
    \centering
    \includegraphics[width=\textwidth]{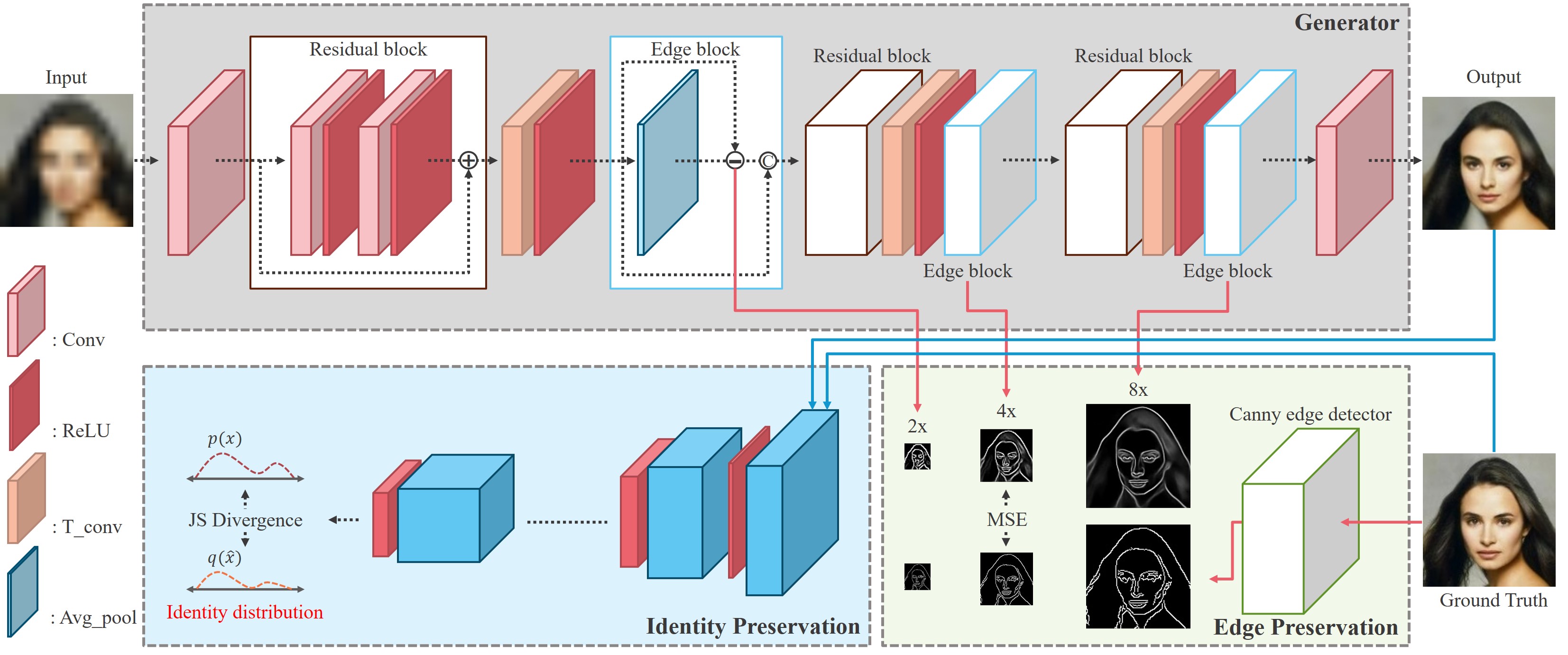}
    \caption{Our proposed network structure. The residual block is comprised of two convolutional layers with ReLU \cite{relu}. The edge block is explained in Section 3.1. $"Conv"$, $"T\_conv"$, and $"Avg\_pool"$ indicate a convolutional layer, a transposed convolutional layer, and a average pooling layer, respectively. We apply the InceptionV1 \cite{inception} network pretrained on VGGFace2 \cite{vggface2} for an identity preservation network.}
    \label{overallnetwork}
\end{figure}

On the other hand, current face SR methods \cite{FSRNet, progressive, exemplar, heatmaps, tiny} have used a facial attention network to emphasize facial landmarks. The attention network highlights the facial components by multiplying attention masks or concatenating attention maps on original feature maps. However, these methods require additional deep neural networks to extract the facial landmarks. Although the landmarks positively affect the reconstruction of face structures, these methods lead to increase in computation cost and runtime. Moreover, the attention network needs extra labels for the facial landmarks to update its trainable parameters, but it is extremely difficult to obtain labels in the real-world.

To solve the aforementioned problems, we propose edge and identity preserving network for face SR, called EIPNet, which elaborately reconstructs facial components in face SR images by concatenating edge information to the original feature maps as shown in Fig. \ref{edge img}. Compared to existing edge detection networks for SR \cite{Edgeinform, SREdgeNet, seanet}, the proposed edge block is super lightweight which consists of a single convolutional layer and a single average pooling layer, but shows high-performance in detecting edge information. We embed the proposed edge block after each transposed convolutional block in multiple scales to provide structural information in each up-scaling process. Furthermore, we define an identity loss function to overcome unlabeled identity classification by comparing class-encoded vectors of SR images and their ground truth. Specifically, an identity network extracts class-encoded vectors of SR images and their ground truth, and the identity loss function applies divergence to two vectors to maximize the similarity. We provide luminance-chrominance error (LCE) which simply converts RGB to YUV space to independently utilize luminance and chrominance information as a loss function. The LCE method reflects distances between SR images and HR images in two domains of RGB and YUV, which provides rich image information to our network. Therefore, our network reconstructs face SR images by accurately restoring facial details. The overall network structure is described in Fig. \ref{overallnetwork}.

We evaluate the performance of our network on CelebA \cite{CelebA} and VGGFace2 \cite{vggface2} datasets. For quantitative measurements, we use Peak Signal to Noise Ratio (PSNR), Structural Similarity (SSIM) \cite{SSIM}, and Visual Information Fidelity (VIF) \cite{sheikh2006image} as evaluation metrics. However, we do not perform subjective evaluations because mean-opinion-score (MOS) that is subjective and difficult to conduct a fair comparison. We use facial region PSNR (FR-PSNR) and facial region SSIM (FR-SSIM) to concentrate on facial details by cropping images with a face detection method \cite{facerecognition}. The FR-PSNR and FR-SSIM remove background and focus on evaluating face regions for a exact and fair comparison.

Compared with existing emthods, our main contributions can be described as follows:
\begin{itemize}
    \item We propose the edge and identity preserving network for face SR to deal with the distortion of facial components by providing the edge information and data distributions. The proposed edge block enables our network to elaborately generate face SR images. Furthermore, the identity loss compares class-encoded vectors to match identities between SR images and their ground truth without identity labels. 
    \item We propose the luminance-chrominance error (LCE) to align global shapes and colors. The LCE method converts RGB to YUV space by using a conversion matrix, which separately generates luminance and chrominance information. We directly utilize these information to reflect distances between face SR images and ground truth with a reconstruction loss of mean-squared-error (MSE).
    \item As evaluation metrics for face SR, we propos FR-PSNR and FR-SSIM to evaluate the restoration results of facial regions. The evaluation metrics remove background components by cropping the output images with a face detection method, and emphasize facial regions for performance comparison.
\end{itemize}

\section{Related Work}

\subsection{Edge Information for Super-resolution}
Intuitively, a single image can be divided into two components: low-frequency and high-frequency parts. The high-frequency component represents detailed information, called as edges, to construct structures of objects. Inspired by the property of edge, the state-of-the-art methods \cite{Edgeinform, SREdgeNet, seanet} have focused on reconstruction of the high-frequency component. In \cite{SREdgeNet}, three kinds of modules were proposed to generate SR images. These modules were composed of SR network, Edge network, and Merge network. The SR network generated rough SR images, which were directly fed to the Edge network for edge extraction. The outputs from the SR network and the Edge network were concatenated together. To reconstruct the high-frequency components in the outputs of the SR network, the Merge network combined the coarse images and their edge maps. Similar to the proposed method in \cite{SREdgeNet}, the edge maps were utilized to enhance SR images in \cite{Edgeinform}. A main contribution of \cite{Edgeinform} was that the SR problem was reformulated as an image inpainting task. Increasing the resolution of a given LR image required recovery of pixel intensities between every two adjacent pixels. The missing pixels were considered as missing regions of the inpainting task. To conduct the inpainting task, the edge maps sketched connection lines between the missing pixels in extended low-resolution images and an SR network carries out image completion. Fang \emph{et al.} \cite{seanet} also employed an edge network to provide edge maps to an SR network. The aforementioned methods enhanced SR images by using edge information; however, an additional deep network is required to extract edge information.

\subsection{Facial Landmarks}
In an upscaling process from low-resolution to high resolution face images, the distortion of facial components is the main obstacle. To address this problem, most state-of-the-art methods \cite{FSRNet, progressive, heatmaps, CAGFace, superfan, tiny} utilized facial landmarks or facial priors. FSRNet \cite{FSRNet} proposed a prior estimation network to extract facial landmark heatmaps. Firstly, this method roughly generated a coarse SR face image and extracted the prior information by using an Hour-Glass (HG) network \cite{hourglass}. Secondly, the prior heatmaps were concatenated with feature maps from an encoder network. Finally, feature maps were fed into a decoder network to generate SR images. In a similar way, Yu \emph{et al.} \cite{heatmaps} also employed the HG network to detect facial landmarks and utilize the information. In \cite{progressive, superfan}, a face alignment network (FAN) was proposed to compare facial heatmaps between SR images and target images. Kim \emph{et al.} \cite{progressive} constructed a single-scale network structure and trained the network to imitate output of the pre-trained HG network by using MSE loss. These tendencies proved that the HG network was an optimized estimator for extracting facial landmarks, but it accompanied high computational cost with labels of facial landmarks. To overcome the dependence on facial landmarks, GWAInet \cite{exemplar} proposed an exemplar guiding method which provided a guiding face to refer its details and preserved the same identity in a restoration process. The method needs extra networks for warping the guiding image to input image and extracting feature maps from the warped image. The network showed comparable results with the restoration of the facial components. However, the network required extra labels (e.g. guiding face images) to train the network and it is difficult to obtain the labels in the real-world.

\section{Proposed Method}

\subsection{Edge Block}

\begin{figure}[t]
    \centering
    \includegraphics[width=8.0cm]{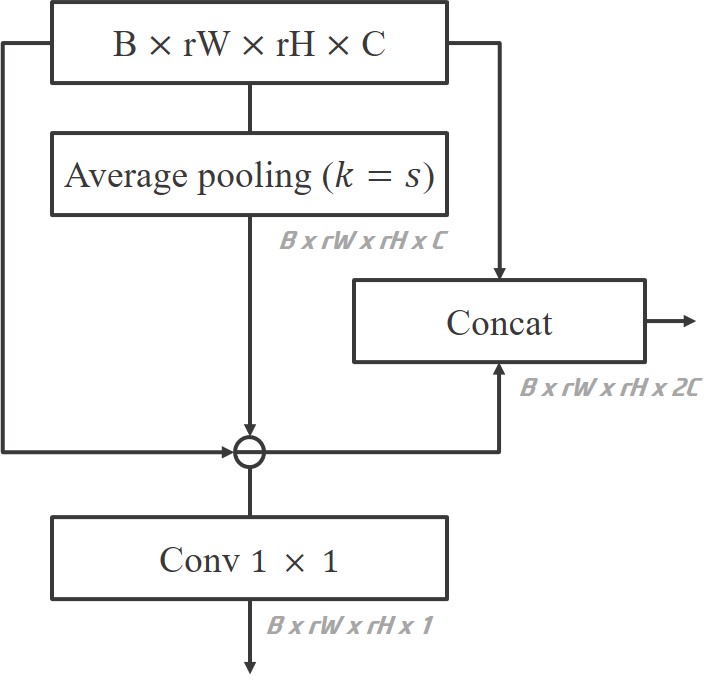}
    \caption{Network structure of the proposed edge block. $r$ is the scaling factor to consider multiple scales and $s$ is the kernel size of the average pooling layer in each scale. $\ominus$ indicates subtraction. The average pooling with padding imitates a low-pass filter to generate smoothened feature maps.}
    \label{edge img}
\end{figure}

Edge information is utilized to enhance images in various tasks such as single image SR \cite{Edgeinform, SREdgeNet, seanet, lee2020epsr}, image inpainting \cite{edgeconnect} and image denoising \cite{denoising}. Without any doubt, edge information is a critical factor to not only elaborately reconstruct facial details from low-resolution face images but also solve an inherent problem of $l_{2}$ loss which tends to generate smoothed images in SR. In detail, the edge information compensates for missing parts in high-frequency components, such as edges and shapes, to enhance the smoothed images. To extract and utilize edge information from features, we propose a lightweight edge block that is comprised of a single convolutional layer and a single average pooling layer as shown in Fig. \ref{edge img}. First, the edge block conducts an average pooling layer like a low-pass filter with different kernel sizes ($k=s$)  in each embedding scale of the edge block to generate smoothed feature maps. Second, the edge block obtains high-frequency components by subtracting the smoothed feature maps from the original feature maps. Finally, the subtracted edge features are concatenated with the input of this block to convey the edge information to the next residual block. To update the proposed edge block, a $1\times1$ convolutional layer reduces the number of the subtracted feature maps to a single-channel matched with edge information of high-resolution images generated by the canny edge detector \cite{canny}. We empirically set the thresholds for the canny edge detector to 100 and 255 by applying adaptive threshold selection \cite{threshold}, which uses distributions of the image gradient magnitude. We calculate the image gradient to obtain gradient magnitude by applying $3\times3$ gradient filters to images, and then we extract the mean $m$ of the gradient magnitude and standard deviation $\sigma$ from images. Based on $m$ and $\sigma$, we calculate the higher threshold as follows:
\begin{equation}
T_{h}=m+k*\sigma,
\end{equation}
where $k$ is its coefficient, and $k\in(1.2, 1.6)$. The lower threshold is defined as:
\begin{equation}
T_{l}=T_{h}/2.
\end{equation}
We set $k$ to be 1.2 and 1.6 to obtain $T_{l}$ and $T_{h}$, respectively. Furthermore, we optimize the kernel size $s$ to extract accurate edges by minimizing Euclidean distance between extracted edges and canny edges. We set the kernel size to be (5, 7, 10). The loss function for the edge block is defined as:
\begin{equation}
    L_{e}=\frac{1}{r^{2}wh}\sum_{x=1}^{rw}\sum_{y=1}^{rh}\left \|C(I_{HR})(x,y)-E(I_{LR})(x,y)\right \|^{2},
    \label{edge filter}
\end{equation}
where $E$ is the edge block, $C$ is the canny edge detector, $I_{HR}$ and $I_{LR}$ are target face images and input low-resolution images, respectively. We consider edge information in multiple scales with scaling factor $r$ which is set to be 2, 4, and 8.

\subsection{Identity Loss}
\begin{figure}[t]
\begin{center}
\includegraphics[width=\textwidth]{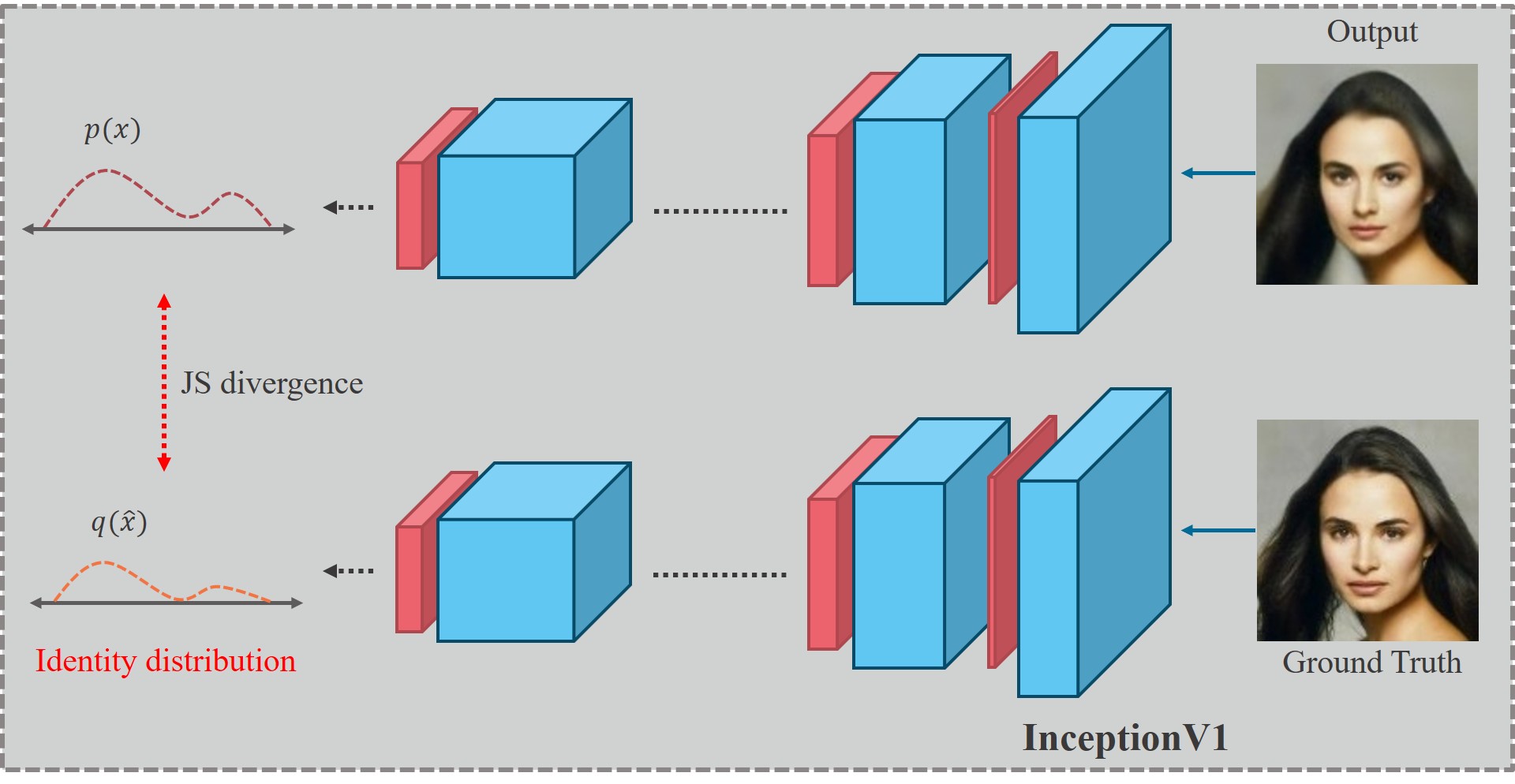}
\caption{Flow chart of the proposed identity loss. We adopt the InceptionV1 \cite{inception} network pretrained on VGGFace2 \cite{vggface2} to extract class-encoded vectors between output and ground truth images.}
\vspace{-0.3cm}
\label{vgg}
\end{center}
\end{figure}

Gottfried Wilhelm Leibniz, who was a German polymath and logician, mentioned no two leaves are alike in a forest of a hundred thousand trees. Also, no two identities are alike in the world. It means that the identity is a unique criterion in distinguishing each people. Even if some methods alleviate the distortion of facial details, it is not meaningful that SR images do not have the same identity as their target images. Inspired by the concept, we propose a novel identity loss to minimize the distance of the identity as shown in Fig. \ref{vgg}. We use Kullback-Leibler (KL) divergence including cross-entropy to calculate the difference of two identities. The relationship between KL divergence and cross-entropy is described as follows:
\begin{equation}
\begin{split}
        H_{p,q}(X)&=-\sum_{i=1}^{N}p(x_{i})\log q(\hat{x_{i}}) \\
                  &=D_{KL}(p\parallel q) + H_{p}(X),
\end{split}
\end{equation}
where $x$ and $\hat{x}$ are ground truth and SR images respectively, $p$ and $q$ are the probability distributions of the identity vector. In addition, $D_{KL}(p\parallel q)$ can be summarized as $\sum_{i=1}^{N}p(x_{i})(\log p(x_{i})-\log q(\hat{x_{i}}))$. Intuitively, this equation is able to obtain the difference between two distributions while cross-entropy is indirect calculation to obtain the difference since entropy of $p(x)$ is discarded. Thereby, KL divergence can be considered as a loss function to infer the difference between two distributions. However, $D_{KL}(p\parallel q)$ yields an infinite value if $p(x)$ is null. Moreover, it has an asymmetric property when the position of distribution $p(x)$ and $q(\hat{x})$ is changed, which precludes exact distance calculation. To solve the problems, we adopt Jensen-Shannon (JS) divergence as the identity loss function which can be described as follows:
\begin{equation}
    D_{JS}=\frac{1}{2}D_{KL}(p\parallel \frac{p+q}{2})+\frac{1}{2}D_{KL}(q\parallel \frac{p+q}{2}).
\end{equation}
We extract 512-bins from targets and SR images to generate class-encoded vectors by using the InceptionV1 \cite{inception} network pretrained on VGGFace2 dataset \cite{vggface2}. The InceptionV1 model is suitable to extract identity properties as the model is trained to categorize the large-scale identities. Firstly, the class-encoded vectors are normalized to $[0, 1]$ by a softmax function. Then, the JS divergence is directly applied to the normalized vectors to calculate the distance between two distributions. As a result, the final loss function for identity is:
\begin{equation}
\begin{split}
    L_{id}=&\frac{1}{2}\sum_{i=1}^{512}\hat{p}(x_{i})(\log \hat{p}(x_{i})-\log (\frac{\hat{p}(x_{i})+\hat{q}(\hat{x_{i}})}{2})) \\
    &+\frac{1}{2}\sum_{i=1}^{512}\hat{q}(\hat{x_{i}})(\log \hat{q}(\hat{x_{i}})-\log (\frac{\hat{p}(x_{i})+\hat{q}(\hat{x_{i}})}{2}))
\end{split}
\end{equation}

where $\hat{p}(x_{i})$ and $\hat{q}(\hat{x_{i}})$ are the class-encoded vectors of ground truth and SR images respectively.

\subsection{Luminance-Chrominance Error}
RGB domain has been adopted as the most popular image representation format by researchers. In SR areas, most state-of-the-art methods also utilized the RGB domain with mean-square-error (MSE) loss to compare between their outputs and ground truth. However, based on experimental results in \cite{yuvvsrgb}, the YUV representation of an image has perceptually better quality than RGB. Furthermore, RGB channels contain blended information of luminance and chrominance components, which causes redundancy of channel information, color distortion, and color shift.

To generate high quality SR images while alleviating the color distortion, we propose a luminance-chrominance error loss which not only converts RGB to YUV image space to split brightness and color components independently but also provides an additional criterion to update the proposed network for the pixel-to-pixel task.

To be specific, we apply YUV conversion matrix to RGB channels to split luminance and chrominance as follows:
\begin{equation}
    \begin{bmatrix}  
Y\\ 
U\\ 
V
\end{bmatrix}=\begin{bmatrix}
0.299 & 0.587 & 0.114\\ 
-0.14713 & -0.28886 & 0.436\\ 
0.615 & -0.51499 & -0.10001
\end{bmatrix}\begin{bmatrix}
R\\ 
G\\ 
B
\end{bmatrix}
\label{TUV matrix}
\end{equation}
where $Y$ represents luminance information, $U$ and $V$ represent chrominance information.

We apply the conversion matrix to target and generated images and compute MSE of the converted two sets of images as follows:  
\begin{equation}
    L_{lc}=\frac{1}{wh}\sum_{c=1}^{yuv}\sum_{x=1}^{w}\sum_{y=1}^{h}\left \|M(I_{HR})(x,y)-M(G(I_{LR}))(x,y)\right \|^{2}
    \label{lc loss}
\end{equation}
where $M$ is the YUV conversion operator, $G$ is our generator network, $c$ is channels of YUV.

\subsection{Overall Loss Function}
In addition to the loss functions discussed so far, we apply a couple of more loss functions as follows:

Context loss: We use a pixel-wise MSE loss to minimize the distances between target images and generated images. The MSE loss is defined as:
\begin{equation}
    L_{rgb}=\frac{1}{wh}\sum_{c=1}^{rgb}\sum_{x=1}^{w}\sum_{y=1}^{h}\left \|I_{HR}(x,y)-G(I_{LR})(x,y)\right \|^{2}
    \label{rgb loss}
\end{equation}
where $c$ is channels of RGB.

Adversarial loss: We use a Generative Adversarial Network (GAN) method to generate realistic SR face images. The adversarial loss is defined as:
\begin{equation}
    L_{ad}=\mathbb{E}[logD(I_{HR})]-\mathbb{E}[log(1-D(G(I_{LR})))]
    \label{gan loss}
\end{equation}
where $\mathbb{E}$ denotes the expectation of the probability distribution, $D$ is the adversarial network.

Total loss: In our final network, we aggregate all loss functions. Thus, the total loss for training is described as:
\begin{equation}
    L_{total} = L_{rgb}+\gamma L_{e}+L_{lc}+\alpha L_{id}+\beta L_{ad}
\end{equation}
where $\gamma$, $\alpha$, $\lambda$ and $\beta$ are the weighting factors which are set empirically.

\subsection{Network Architecture}
\begin{table*}[t]
\tiny
\centering
\caption{Configuration of the edge and identity preserving network for face super-resolution (EIPNet). All edge information is extracted after each transpose convolutional layer.}
\begin{tabular}{|c|c|c|c|c|c|}
\hline
Layer                                                                                   & \begin{tabular}[c]{@{}c@{}}ReLU activation\end{tabular} & Size of filter & Stride & Output channel & Output size ($h\times w$) \\ \hline \hline
\it{conv0}                                                                              & X                                                                          & $3\times3$     & 1      & 512            & $16\times16$              \\ \hline
\it{conv1\_1}                                                                           & O                                                                          & $3\times3$     & 1      & 512            & $16\times16$              \\ \hline
\it{conv1\_2}                                                                           & O                                                                          & $3\times3$     & 1      & 512            & $16\times16$              \\ \hline
\begin{tabular}[c]{@{}c@{}}\it{residual block1}\\ \it{(conv0+conv1\_2)}\end{tabular}     & -                                                                              & -              & -      & 512            & $16\times16$              \\ \hline
\it{transpose conv1}                                                                    & O                                                                          & $4\times4$     & 2      & 256            & $32\times32$              \\ \hline
\begin{tabular}[c]{@{}c@{}}\it{edge block1}\\ \it{(t\_conv1, edge1)}\end{tabular} & -                                                                              & -              & -      & 512            & $32\times32$              \\ \hline
\it{conv2\_1}                                                                           & O                                                                          & $3\times3$     & 1      & 512            & $32\times32$              \\ \hline
\it{conv2\_2}                                                                           &  O                                                                          & $3\times3$     & 1      & 256            & $32\times32$              \\ \hline
\begin{tabular}[c]{@{}c@{}}\it{residual block2}\\ \it{(t\_conv1+conv2\_2)}\end{tabular}  & -                                                                              & -              & -      & 256            & $32\times32$              \\ \hline
\it{transpose conv2}                                                                    &  O                                                                          & $4\times4$     & 2      & 128            & $64\times64$              \\ \hline
\begin{tabular}[c]{@{}c@{}}\it{edge block2}\\ \it{(t\_conv2, edge2)}\end{tabular} & -                                                                              & -              & -      & 256            & $64\times64$              \\ \hline
\it{conv3\_1}                                                                           &  O                                                                          & $3\times3$     & 1      & 256            & $64\times64$              \\ \hline
\it{conv3\_2}                                                                           & O                                                                          & $3\times3$     & 1      & 128            & $64\times64$              \\ \hline
\begin{tabular}[c]{@{}c@{}}\it{residual block3}\\ \it{(t\_conv2+conv3\_2)}\end{tabular}  & -                                                                              & -              & -      & 128            & $64\times64$              \\ \hline
\it{transpose conv3}                                                                    & O                                                                          & $4\times4$     & 2      & 64             & $128\times128$            \\ \hline
\begin{tabular}[c]{@{}c@{}}\it{edge block3}\\ \it{(t\_conv3, edge3)}\end{tabular} & -                                                                              & -              & -      & 128            & $128\times128$            \\ \hline
\it{conv4}                                                                              &  X                                                                          & $3\times3$     & 1      & 3              & $128\times128$            \\ \hline
\end{tabular}
\label{table1}
\end{table*}

\begin{table}
\centering
\caption{Configuration of the discriminator. Without the last fully-connected layer, Leaky ReLU with $\alpha=0.2$ is applied after each layer.}
\begin{tabular}{|c|c|c|c|}
\hline
Layer    & Size of filter & Stride & Output channel \\ \hline \hline
\it{conv1\_1} & $3\times3$     & 1      & 128            \\ \hline
\it{conv1\_2} & $3\times3$     & 2      & 128            \\ \hline
\it{conv2\_1} & $3\times3$     & 1      & 256            \\ \hline
\it{conv2\_2} & $3\times3$     & 2      & 256            \\ \hline
\it{conv3\_1} & $3\times3$     & 1      & 256            \\ \hline
\it{conv3\_2} & $3\times3$     & 2      & 512            \\ \hline
\it{conv4}    & $3\times3$     & 1      & 512            \\ \hline
\it{fc}       & -              & -      & 512            \\ \hline
\it{fc}       & -              & -      & 1              \\ \hline
\end{tabular}
\label{discriminator}
\end{table}

The proposed network architecture in Fig. \ref{overallnetwork} is comprised of three residual blocks, three transposed convolutional blocks, and two convolutional layers in the first and last position of our network. Furthermore, three edge blocks are embedded after each transposed convolutional block. The overview on the configuration of our network is summarized in Table \ref{table1}. Specifically, the residual block consists of two $3\times3$ convolutional layers with Rectifier Linear Unit (ReLU) as the activation function. The transposed convolutional block also uses ReLU the same as the residual block. Each transposed block conducts $2\times$ upscaling and the final output is $8\times$ upscaled face images from $16\times16$ low-resolution input. After upsampling feature maps, the edge block is utilized to obtain the sharp edge information on facial components. The edge block has a single $1\times1$ convolutional layer and a single average pooling layer. After finishing all the processing, the last $3\times3$ convolutional layer converts feature maps to RGB channels.

To generate realistic SR images, we adopt the adversarial loss function. Concretely, our discriminator network is comprised of seven convolutional layers with Leaky ReLU and two fully-connected layers as described in Table \ref{discriminator}.

\section{Experiments}

\subsection{Datasets}
To evaluate our methods, we use two challenging datasets for face SR: CelebA \cite{CelebA} and VGGFace2 \cite{vggface2}. The CelebA dataset is a large-scale face dataset with about 0.2 million images. The dataset provides aligned and unaligned face images. The aligned face images are cropped into squares. We use both of face images to train and test our model. For aligned face images, we use 162,770 images as a training set and 19,962 images as a test set. Furthermore, we also evaluate our model using unaligned face images. The unaligned dataset is comprised of 135,516 train images from a part of a test set in the VGGFace2 dataset and 33,880 test images from the remaining of the test set.

\subsection{Implementation Details}
We perform data augmentation as follows. We conduct center cropping with $178\times178$ resolution and resizing $128\times128$ for the CelebA dataset. For the VGGFace2 dataset, we crop images with $0.7\times$ minimum resolution between height and width and resize $128\times128$. After the resizing process, we horizontally flip input images with a probability of 0.5 and rotate the images by $90^{\circ}$ and $270^{\circ}$. We progressively perform $2\times$ downscaling to generate $8\times$ downscaled low-resolution input images. As a test manner, we only use cropping and resizing by using the same resolution as that in training, and also apply $8\times$ downsampling for test images by using bilinear interpolation. For training, we use Adam optimizer \cite{Adam} with $\beta_{1}=0.9$, $\beta_{2}=0.999$ and $\epsilon=1e-8$ in the generator network, and $\beta_{1}=0.5$ and $\beta_{2}=0.9$ in the discriminator network. The learning rate is set to be $1e-4$.

We train each dataset in the same conditions except the adversarial loss parameter as $\beta=0$ until the iteration reaches 1.2M. Then, we train our networks with the adversarial loss function until the iteration reaches 1.32M. We conduct experiments on a single Tesla V100 GPU for training and a single GTX 1080ti for evaluation. All codes are written and tested in Python with TensorFlow \cite{tensorflow}. We will release the codes in github\footnote{\url{https://github.com/BenjaminJonghyun/EIPNet}.}

\subsection{Evaluation Metrics}
We use Peak Signal to Noise Ratio (PSNR), Structural Similarity (SSIM), Visual Information Fidelity (VIF), Facial Region PSNR (FR-PSNR), and Facial Region SSIN (FR-SSIM) to quantitatively evaluate our network. Specifically, the FR-PSNR and FR-SSIM are used to replace subjective evaluation metrics, i.e., Mean-Opinion-Score (MOS). To calculate FR-PSNR and FR-SSIM, we crop the face region from an HR image detected by a face recognition method \cite{facerecognition} without background. The cropped area is applied to its SR counterpart for matching with the face region of the HR image. Then, we calculate PSNR and SSIM between two cropped images. In this process, if the face recognition method fails to detect the face region in the HR image, this image is discarded in the test set. Therefore, 18,807 of 19,962 and 31,779 of 33,880 images are tested for FR-PSNR and FR-SSIM on CelebA and VGGFace2, respectively.

\subsection{Ablation study}

\subsubsection{Edge Block}
\begin{figure*}[t]
    \centering
    \includegraphics[width=\textwidth]{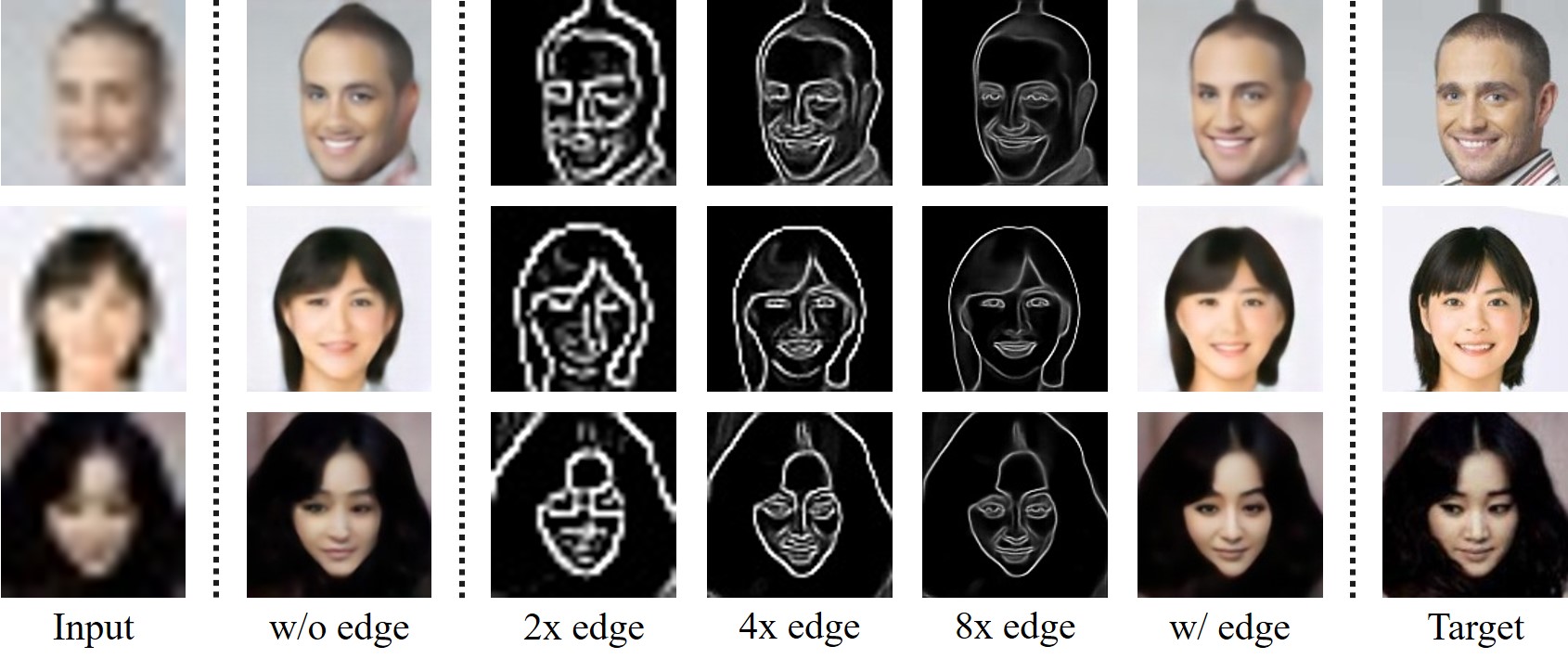}
    \caption{Comparison of super-resolved images with/without the edge block. The input image has $16\times16$ resolution and other outputs have $128\times128$ resolution. The edge images are progressively extracted from each edge block. We resize the input and edge images to $128\times128$ to intuitively show the figures.}
    \label{comp_edge}
\end{figure*}

Our SR network employs the edge block in each upscaling process to preserve the high-frequency component. To verify that the proposed edge block preserves edges and boundaries of facial components, we conduct experiments with/without the edge block. Specifically, Fig. \ref{comp_edge} shows that the visibility of eyes and mouth is improved by embedding the edge block compared to outputs of the baseline network which use only $L_{rgb}$ loss function. As can be seen from Fig. \ref{comp_edge}, the outputs with the edge block are clearly improved over those without the edge block. Furthermore, Fig. \ref{comp_edge} shows that the edge block extracts sharp edge information of facial components in multiple scales by simply subtracting blurred feature maps from the original feature maps. The subtracted feature maps are concatenated to the original feature maps to complement high-frequency components. Such aggregation guides the EIPNet to elaborately reconstruct high-frequency components in facial components and generate high quality SR images. Furthermore, we provide experiments on CelebA to analyze the effects of the edge block embedding in different scales. In Table  \ref{edgescale}, it is obvious that the edge block embedding in multiple scales improves the network performance in both PSNR and SSIM. Specifically, mid-level aggregation ($4\times$) shows the largest improvement among single-scale embedding since the mid-level edges contain both global and local high-frequency components simultaneously. In the multi-scale embedding, the combination of three scales achieves the best performance.

\begin{table}[t]
\centering
\caption{Effects of edge block embedding in different scales on the performance. We use CelebA dataset for evaluation.}
\begin{tabular}{c|c|c||c||c}
\hline
\multicolumn{3}{c||}{Scaling factor}                           & \multirow{2}{*}{PSNR}      & \multirow{2}{*}{SSIM}       \\ \cline{1-3}
$2\times$    & $4\times$    & $8\times$             &                            &                             \\ \hline \hline
\multicolumn{3}{c||}{Baseline ($L_{rgb}$)} & 23.11 & 0.7070 \\ \hline \hline
$\checkmark$ &              &                       & 23.87                      & 0.7169                      \\ \hline
             & $\checkmark$ &                       & 24.06                      & 0.7193                      \\ \hline
             &              & $\checkmark$          & 23.99                      & 0.7181                      \\ \hline
             & $\checkmark$ & $\checkmark$          & 24.52                      & 0.7297                      \\ \hline
$\checkmark$ & $\checkmark$ &                       & 24.31                      & 0.7244                        \\ \hline
$\checkmark$ & $\checkmark$ & $\checkmark$          & \textbf{24.74}                      & \textbf{0.7339}                      \\ \hline
\end{tabular}
\label{edgescale}
\vspace{-0.5cm}
\end{table}

The results prove that our edge network is effective in preserving edge information in the face SR under a shallow and simple network. In order to reconstruct or enhance images using a convolutional layer, the experimental results imply that our edge block can be simply applied to other domains with a marginal increase in learnable parameters. However, the proposed method regards noise as edge components (See the results of the edge block). The reason is that the subtraction operation for extracting the high-frequency components preserves high-value pixels including edges, contours, and noises. To solve it, the residual block is embedded after the edge block to refine noisy edges and combine them with feature maps

\begin{figure}[t]
    \centering
    \includegraphics[width=9.0cm]{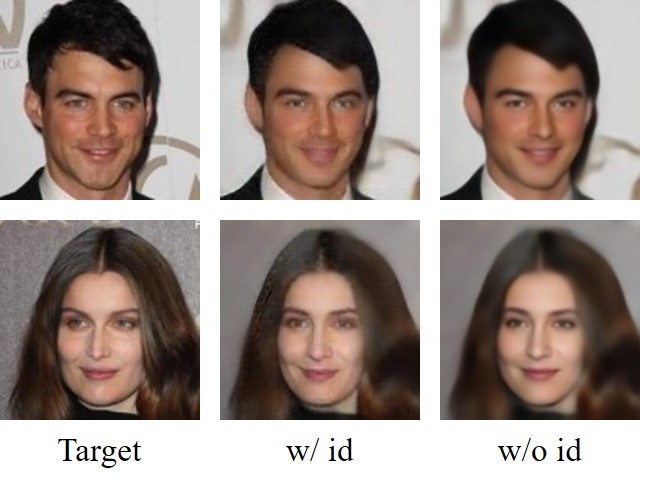}
    \caption{Comparison of super-resolved images with/without the identity loss. The examples are True and False cases compared to identities of the target images.}
    \label{idexample}
\end{figure}

\subsubsection{Identity Loss}
To prove the positive effect of the proposed identity loss, we evaluate SR images with/without the identity loss in terms of preservation of identities. This experiment is conducted by a face recognition algorithm \cite{facerecognition} and a face recognition network \cite{vggface2} to compare identities between target and SR images. Fig. \ref{idexample} is true and false cases of the experiment which shows that the outputs with the identity loss are a true case while the outputs without the identity loss are a false case. It can be observed that the identity loss reconstructs characteristics of facial details to make them perceptually similar to the target images. 

In order to quantify the identity comparison, we use a True Acceptance Rate (TAR) evaluation metric as follows:

\begin{equation}
    TAR=\frac{N_{TA}}{N_{ALL}}\times 100,
\end{equation}
where $N_{TA}$ is the number of true-accept images, which is determined by a distance threshold $d$ between vectors extracted from an SR image and its ground truth, and $N_{ALL}$ is the number of total images in the test set. For evaluation of the proposed method in TAR based on the face recognition algorithm, the distance is calculated by Euclidean distance while the squared $L_{2}$ distance is applied to network based evaluation.

According to the testing rule, our method with the identity loss shows $64\%$ accuracy which is about $6\%$ higher than the exclusion method. The experimental result proves that our identity loss is effective in preserving identities in the upscaling process. Furthermore, the proposed method alleviates inherent problems in the face SR such as blurring effect and unnaturalness. 

\subsubsection{Luminance-Chrominance Error}
\begin{figure}
\centering
\includegraphics[width=9.0cm]{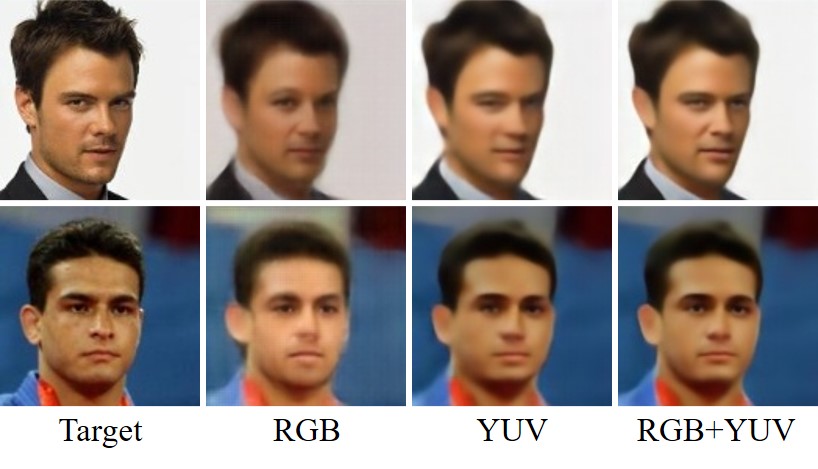}
\caption{Comparison of super-resolved images with $L_{rgb}$ (RGB) and  $L_{lc}$ (YUV). }
\label{compyuv}
\end{figure}

We evaluate the effect of luminance-chrominance error (LCE) in terms of quantitative evaluation metrics. We conduct experiments on CelebA dataset to explore the effect of a Luminance-Chrominance Error (LCE) loss function.  Our baseline network is employed for this implementation. As shown in Table \ref{effectyuv}, the YUV loss achieves higher accuracy in both PSNR and SSIM than the RGB loss. Moreover, in Fig. \ref{compyuv}, the LCE loss facilitates the proposed network to generate correct luminance and color information in SR images by separately reconstructing luminance and chrominance components compared with the RGB loss. When we combine RGB loss and YUV loss, the proposed network yields elaborate SR images by fully utilizing pixel information in multi-domains.

The experiment results prove that the luminance-chrominance loss conducts global fine-tune in each pixel by considering luminance and chrominance information separately.

\begin{table}
\centering
\caption{Effects of YUV representation on the CelebA super-resolution task. RGB: $L_{rgb}$ loss. YUV: $L_{lc}$ loss. Two loss functions are adopted to train our baseline.}
\begin{tabular}{|c||c|c|c|}
\hline
Loss & \multicolumn{1}{l|}{RGB} & \multicolumn{1}{l|}{YUV} & \multicolumn{1}{l|}{RGB+YUV} \\ \hline \hline
PSNR & 23.11                        & 23.79                    & 23.83                            \\ \hline 
SSIM & 0.7070                       & 0.7186                   & 0.7204                           \\ \hline
\end{tabular}
\label{effectyuv}
\end{table}

\begin{figure}
    \centering
    \includegraphics[width=8.5cm]{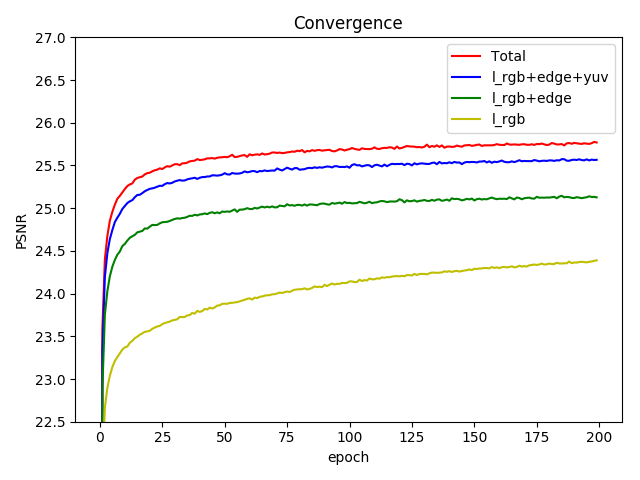}
    \caption{Curves of the training procedure on CelebA for each loss function. We progressively add the proposed loss functions to $L_{rgb}$. The adversarial loss is excluded because a training method with the adversarial loss is different from other loss functions.}
    \label{convergence}
\end{figure}

\begin{figure*}[t]
    \centering
    \includegraphics[width=\textwidth]{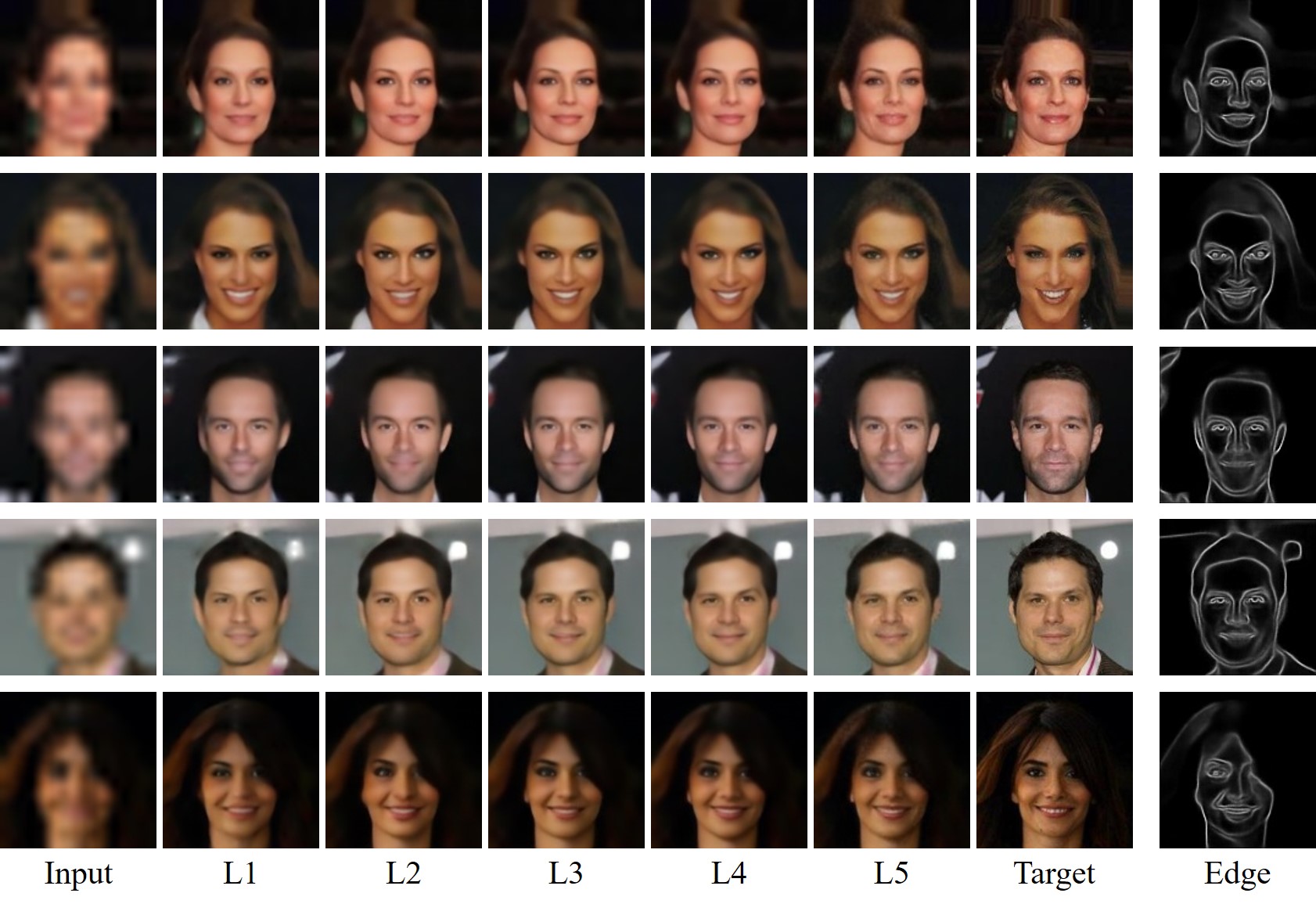}
    \caption{Comparison of our outputs depending on loss functions. The input image has $16\times16$ resolution and other outputs have $128\times128$ resolution. The edge images are extracted from the third edge block. The $L1$ utilizes only $L_{rgb}$. Starting from the $L1$, we progressively add $L_{e}, L_{lc}, L_{id}$ and, $L_{ad}$ to $L5$.}
    \label{overall_output}
\end{figure*}

\subsubsection{Summary}
In this work, we provide three main contributions (edge block, identity loss, and LCE) to enhance SR images. To demonstrate the effectiveness of the three contributions in the training procedure, we provide Fig. \ref{convergence} to represent curves of PSNR values per each epoch on the CelebA dataset for each loss function. These curves show that the three contributions enables the proposed network to converge fast while the $L_{rgb}$ loss gradually converges. They indicate that the three contributions are effective to rapidly reach a peak value in the training procedure.

To prove that the three contributions are also effective in the testing procedure, we provide Fig. \ref{overall_output} and Table \ref{table2} to show progressive improvements of SR images depending on each loss function. In the testing procedure, our network takes $16\times16$ low-resolution images as input and conducts the $8\times$ upscaling process (See the implementation details).

\begin{table*}[t]
\centering
\caption{Ablation study on the identity loss in terms of PSNR, FR-PSNR, SSIM, and FR-SSIM. We use CelebA dataset for evaluation.}
\begin{tabular}{c||cc||cc}
\hline
Method & PSNR & FR-PSNR & SSIM & FR-SSIM \\ \hline \hline
Bicubic & 18.45 & 18.12 & 0.4953 & 0.4565 \\
$L_{rgb}$ & 23.11 & 23.42 & 0.7070 & 0.7278 \\
$L_{rgb}+L_{e}$ & 24.74 & 24.73 & 0.7339 & 0.7522 \\
$L_{rgb}+L_{e}+L_{lc}$ & 25.05 & 25.15 & 0.7461 & 0.7699 \\
$L_{rgb}+L_{e}+L_{lc}+L_{id}$ & 25.16 & 25.31 & 0.7494 & 0.7745 \\
$L_{rgb}+L_{e}+L_{lc}+L_{id}+L_{ad}$ & 25.08 & 25.20 & 0.7429 & 0.7696 \\
\hline
\end{tabular}
\label{table2}
\end{table*}

According to the Fig. \ref{overall_output} and the Table \ref{table2}, our edge block shows a remarkable improvement in both visual quality and quantitative measurements by compensating for missing parts of the high-frequency component in upscaling layers. Specifically, the edge block alleviates blurring effect and shows a large improvement in PSNR and SSIM. On the other hand, identity loss and LCE do not show much difference in visual quality while improving quantitative results. Finally, our method achieves $25.16$ dB in PSNR and $0.7494$ in SSIM. However, the proposed network also suffers from the perception-distortion trade-off as mentioned in \cite{tradeoff}. The adversarial loss decreases the quantitative performance even though the visual quality looks better.

\begin{figure*}[t]
    \centering
    \includegraphics[width=\textwidth]{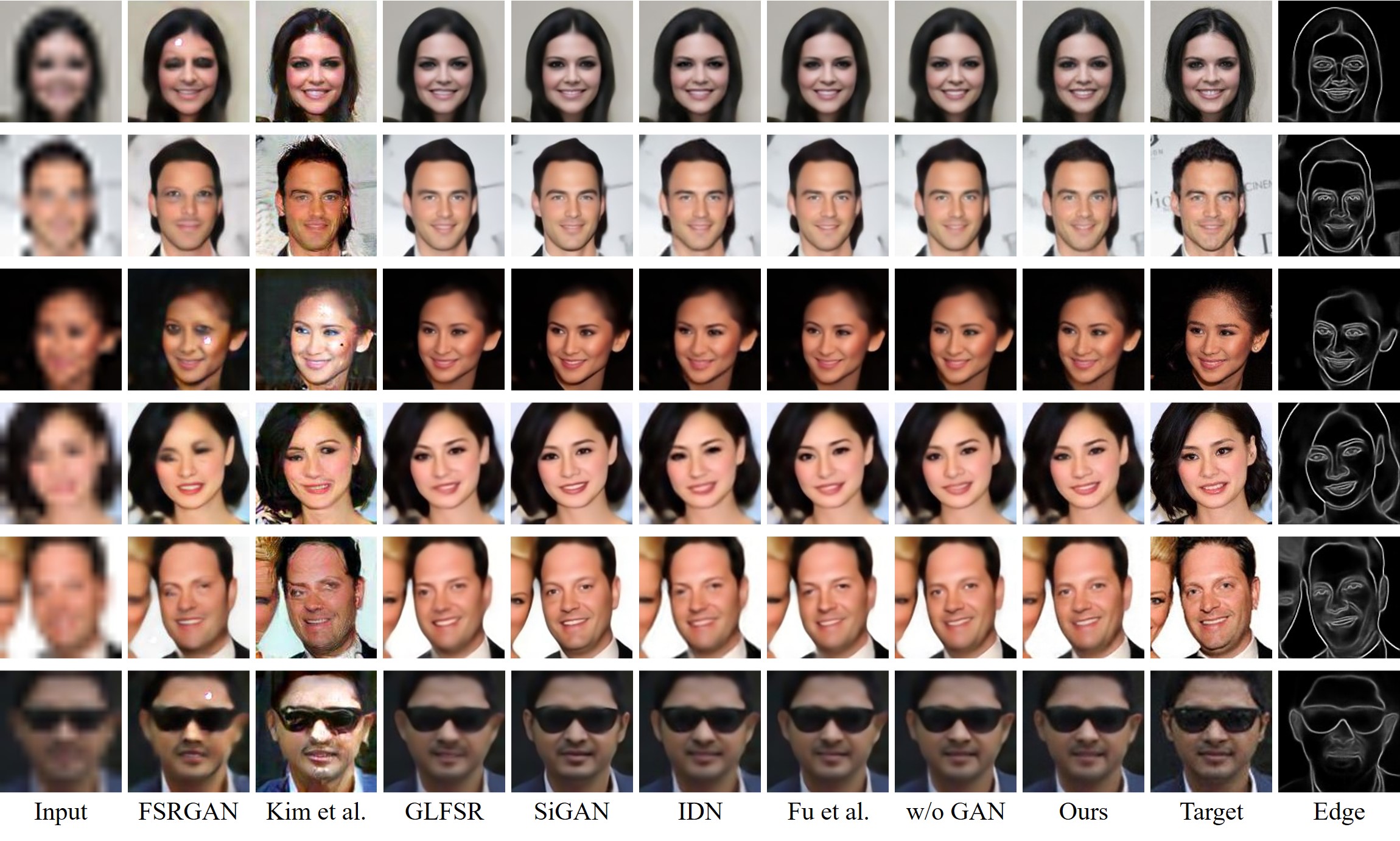}
    \caption{Qualitative comparison with state-of-the-art methods. The size of input and output is the $16\times16$ and $128\times128$ respectively. From the first row to the third row, example images are from CelebA, others are from VGGFace2. In the last column, edge images are generated by our edge block. For the best view, please zoom in this figure.}
    \label{compsota}
\end{figure*}

\begin{table*}[h!]
\small
\centering
\caption{Quantitative comparison with state-of-the-art methods on CelebA and VGGFace2. To concentrate on facial regions, we use FR-PSNR and FR-SSIM criteria. For these criteria, we extract the facial regions from each super-resolved image by utilizing the face recognition method. The best results are \textbf{highlighted.}}
\begin{tabular}{c||l||cc||cc||c}
\hline
Dataset                   & \multicolumn{1}{c||}{Method} & PSNR  & FR-PSNR & SSIM   & FR-SSIM & VIF \\ \hline \hline
\multirow{8}{*}{CelebA}   & FSRGAN \cite{FSRNet}          & 22.84 & 22.13   & 0.6665 & 0.6451 & 0.823  \\ \cline{2-7} 
                          & Kim \emph{et al.} \cite{progressive}  & 23.19 & 23.62   & 0.6798 & 0.7288 & 0.823  \\ \cline{2-7}
                          & GLFSR \cite{GLFSR}            & 24.19 & 24.22   & 0.7136 & 0.7391 & 0.903  \\ \cline{2-7}
                          & SiGAN \cite{SiGAN}            & 24.17 & 24.61   & 0.7255 & 0.7639 & 0.902  \\ \cline{2-7}
                          & IDN \cite{IDN}                & 24.28 & 24.37   & 0.7170 & 0.7424 & \textbf{0.907}  \\ \cline{2-7} 
                          & Fu \emph{et al.} \cite{realtime}      & 24.41 & 24.62   & 0.7250 & 0.7561 & \textbf{0.907}  \\ \cline{2-7}
                          & w/o GAN                     & \textbf{25.16} & \textbf{25.31} & \textbf{0.7494} & \textbf{0.7745} & 0.906 \\ \cline{2-7}
                          & Ours                        & 25.08 & 25.20   & 0.7429 & 0.7696 & 0.901  \\ \hline \hline
\multirow{8}{*}{VGGFace2} & FSRGAN\cite{FSRNet}           & 22.02 & 21.58   & 0.6463 & 0.6423 & 0.819  \\ \cline{2-7} 
                          & Kim \emph{et al.} \cite{progressive}  & 22.65 & 22.23   & 0.6699 & 0.6560 & 0.817  \\ \cline{2-7}
                          & GLFSR \cite{GLFSR}            & 24.06 & 23.78   & 0.6887 & 0.6913 & 0.907  \\ \cline{2-7}
                          & SiGAN \cite{SiGAN}            & 24.29 & 24.67   & 0.7103 & 0.7137 & 0.908  \\ \cline{2-7}
                          & IDN \cite{IDN}                & 24.26 & 24.49   & 0.6969 & 0.6997 & 0.903  \\ \cline{2-7} 
                          & Fu \emph{et al.}\cite{realtime}      & 24.30 & 24.62   & 0.7017 & 0.7083 & 0.905  \\ \cline{2-7} 
                          & w/o GAN                     & \textbf{24.72} & \textbf{25.01} & \textbf{0.7148} & \textbf{0.7201} & \textbf{0.910} \\ \cline{2-7}
                          & Ours                        & 24.56 & 24.83   & 0.7066 & 0.7138 & 0.908  \\ \hline
\end{tabular}
\label{compsotatable}
\end{table*}

\subsection{Comparison with State-of-the-Art Methods}

Each face SR method can be individually categorized by datasets, the size of images, and the ratio of scaling. We compare our method with six stat-of-the-art methods: FSRGAN \cite{FSRNet}, Kim \emph{et al.} \cite{progressive}, GLFSR \cite{GLFSR}, SiGAN \cite{SiGAN}, IDN \cite{IDN}, and Fu \emph{et al.} \cite{realtime}. These methods have the similar experimental conditions to ours to quantitatively and qualitatively compare SR results. Furthermore, we compare the proposed method with these state-of-the-arts in terms of computational cost and runtime.

\subsubsection{Quantitative and Qualitative Comparison}

\begin{table}[t]
\centering
\caption{Comparison with state-of-the-art methods in matching identities on CelebA. This experiment is conducted according to the face recognition testing rule. The face recognition algorithm and network are applied to compare the outputs of each method and ground truth. $d$ is the distance threshold. The best results are \textbf{highlighted.}}
\begin{tabular}{l||c|c|c|c}
\hline
\multicolumn{1}{c||}{\multirow{2}{*}{Method}} & \multicolumn{2}{c|}{Algorithm} & \multicolumn{2}{c}{Network}  \\ \cline{2-5}
\multicolumn{1}{l||}{}        & $d=0.5$        & $d=0.6$       & $d=0.02$      & $d=0.04$      \\ \hline \hline
GLFSR \cite{GLFSR}                          & 5.3            & 43.7          & 19.6          & 52.3          \\ \hline
IDN \cite{IDN}                            & 6.9            & 48.3          & 24.0          & 59.1          \\ \hline
Fu \emph{et al.} \cite{realtime}                      & 11.3           & 59.8          & 24.6          & 58.3          \\ \hline
SiGAN \cite{SiGAN}                          & 14.7           & 62.8          & \textbf{28.7} & \textbf{63.2} \\ \hline
Ours w/o identity loss         & 10.8           & 58.1          & 23.3          & 59.5          \\ \hline
\textbf{Ours w/ identity loss} & \textbf{15.1}  & \textbf{64.0} & 25.2          & 63.0          \\ \hline
\end{tabular}
\label{identity}
\end{table}

\begin{table}[!h]
\centering
\caption{No-reference image assessment scores by the BRISQUE method on CelebA. A lower score is better quality. The best result is \textbf{highlighted.}}
\begin{tabular}{l|l||c}
\hline
\multicolumn{2}{l||}{Method}       & Score \\ \hline \hline
\multicolumn{2}{l||}{Ground-truth} & 35.78 \\ \hline \hline
\multicolumn{2}{l||}{GLFSR}        & 50.90 \\ \hline
\multicolumn{2}{l||}{Fu \emph{et al.}}    & 48.88 \\ \hline
\multicolumn{2}{l||}{IDN}          & 47.50 \\ \hline
\multicolumn{2}{l||}{SiGAN}        & 47.32 \\ \hline
\multicolumn{2}{l||}{FSRGAN}       & 46.96 \\ \hline
\multicolumn{2}{l||}{Kim \emph{et al.}}   & 39.94 \\ \hline
\multirow{2}{*}{Ours}   & w/o GAN  & 48.32 \\ \cline{2-3} 
                        & w/ GAN   & \textbf{32.55} \\ \hline
\end{tabular}
\label{brisque}
\end{table}

To validate the effectiveness of the proposed method, we compare our method with the aforementioned state-of-the-art methods in terms of quantitative and qualitative measures. Table \ref{compsotatable} shows quantitative measurements on different datasets. Experimental results show that the proposed method achieves the best PSNR and SSIM performance on both datasets without GAN. Specifically, the PSNR and SSIM values of the proposed method without GAN are 0.75 dB and 0.0244 higher than Fu \emph{et al.}, which shows the best performance among existing methods with a lightweight structure. Furthermore, the proposed method is effective in preserving visual information while showing the best and comparable performance in VIF on two datasets. However, a higher information fidelity does not mean that the face images are more realistic and natural-looking, and SR images have the same identity as their corresponding HR ones. The reason is that PSNR and SSIM are capable of representing an image quality gain, but they are biased towards blurred images \cite{tradeoff}. Thus, we  use the face recognition test and NR-IQA (No-Reference Image Quality Assessment). For the face recognition test, we apply the face recognition algorithm \cite{facerecognition} and network \cite{vggface2} with TAR to compare identities between SR images and their ground-truth. As can be seen from Table \ref{identity}, the proposed method outperforms existing face SR and real-time SR ones in terms of the algorithm based testing rule while achieving slightly lower performance on the deep learning based test compared to SiGAN \cite{SiGAN}.  Furthermore, to verify our SR images are realistic and natural-looking, we adopt Blind/Referenceless Image Spatial Quality Evaluator (BRISQUE) \cite{brisque} as NR-IQA.  It can be observed from Table \ref{brisque} that the proposed method achieves the best score in BRISQUE based NR-IQA. The result indicates that the proposed method is able to preserve pixel-wise accuracy while generating realistic and natural-looking SR images.

Moreover, we provide some SR images on both datasets as shown in Fig. \ref{compsota}. It is obvious that our method restores correct facial details while other methods generate failed cases. This is because the edge block progressively reconstructs facial structures by concatenating the high-frequency component to the original feature maps. Thus, our method performs better in restoring facial structures and alleviating the blurring effect. Furthermore, our method with GAN generates more realistic images while other methods produce distortions or non-realistic images. 

\begin{figure}
    \centering
    \includegraphics[width=10cm]{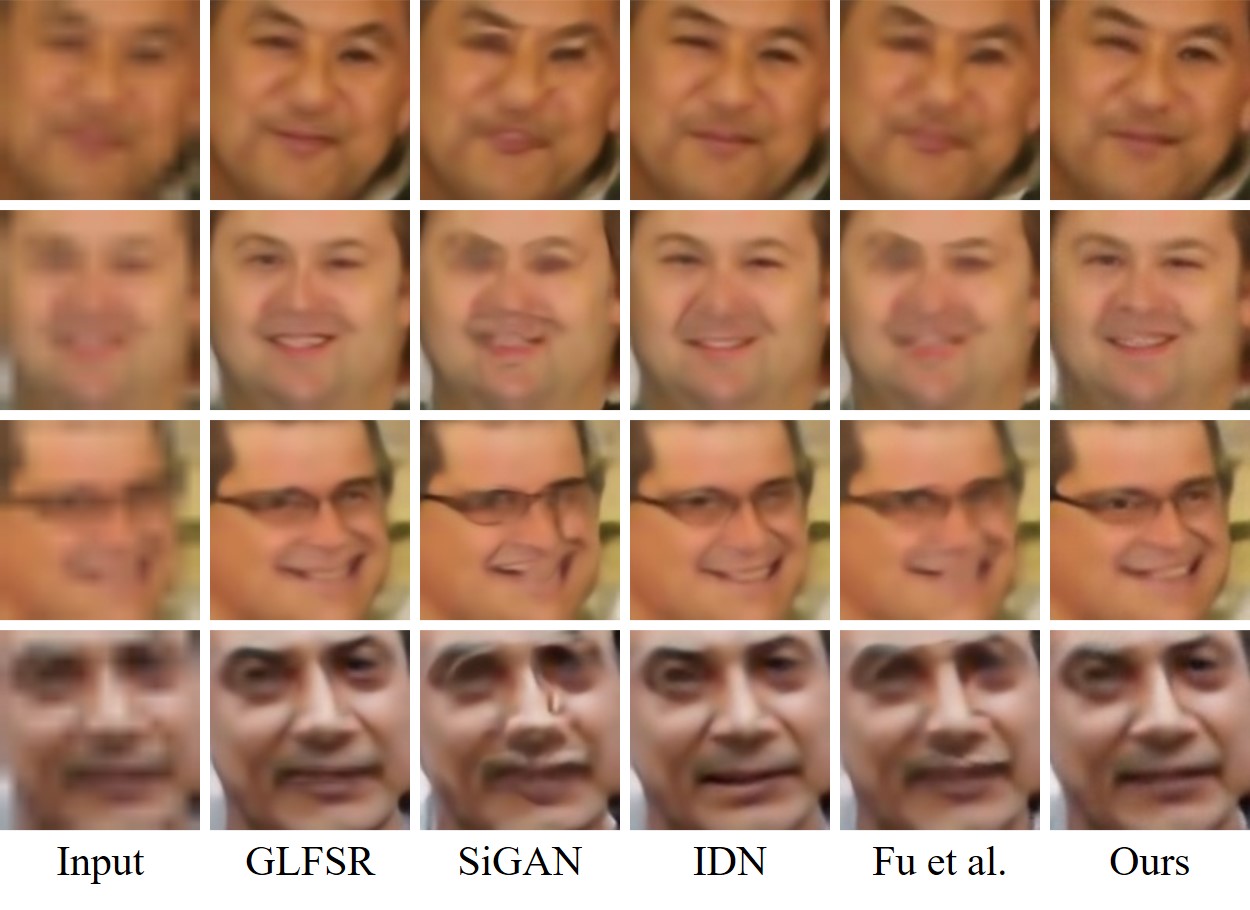}
    \caption{Comparison with state-of-the-art methods on real-world images of WiderFace.}
    \label{real_sota}
\end{figure}

For the real-world application, we compare the proposed method with the existing methods on a WiderFace \cite{yang2016wider} dataset, which is a face detection dataset containing 32,203 images with face bounding boxes. Fig. \ref{real_sota} shows SR results of the existing methods and ours on real LR images from this dataset. It is obvious from this figure that the proposed method is capable of generating consistent SR images while other methods yield distortions in their results.
Consequently, the comparisons on various datasets prove that our method is an effective network to generate outstanding SR images.

\begin{table}[t]
\centering
\caption{Computational complexity comparison with state-of-the-art methods. All the runtime is computed using a single GTX 1080Ti with the Tensorflow platform. The best results are \textbf{highlighted.}}
\begin{tabular}{l||c|c|c|c|c}
\hline
\multicolumn{1}{c||}{Method} & Parameter & GMACs & FPS & PSNR & SSIM \\ \hline \hline
FSRGAN                       & 2.15M     & 15.79  & 62  & 22.84 & 0.6665  \\ \hline
Kim \emph{et al.}                   & 6.83M     & 3.36  & \textbf{317} & 23.19 & 0.6798 \\ \hline
GLFSR                        & 1.61M     & 26.27  & 97  & 24.19 & 0.7136 \\ \hline
SiGAN                        & 156.14M   & 31.08  & 121  & 24.17 & 0.7255 \\ \hline
IDN                          & \textbf{0.68M}     & \textbf{0.17}  & 92 & 24.28 & 0.7170 \\ \hline
Fu \emph{et al.}                    & 2.96M     & 5.54  & 213 & 24.41 & 0.7250 \\ \hline
Ours                         & 11.91M    & 10.16 & 215 & \textbf{25.08} & \textbf{0.7429} \\ \hline
\end{tabular}
\label{comp_comp}
\end{table}

\subsubsection{Computational Comparison}

Since each method employs different devices, we measure computational complexity using a single GTX 1080Ti GPU with the Tensorflow platform for a fair comparison. We evaluate Giga Multiply Accumulate (GMAC) that computes the product of two numbers and adds it to the accumulator. Thus, GMACs indicate computational cost of multiplies and accumulates per second. In this work, we perform the reconstruction of an 128×128 SR face image from a given 16×16 LR face one. Table \ref{comp_comp} shows computational complexity comparison with state-of-the-art methods on CelebA dataset. Kim \emph{et al.} \cite{progressive} outperforms existing methods with a big gain in FPS while quantitative results are lower than the others. On the other hand, IDN \cite{IDN} shows the lowest computational complexity, but the inference time is slower than most state-of-the-art method because this network employs repetition of 4-fragment group convolution. According to experiments in \cite{zhang2018shufflenet}, the group convolution shows the 3 times slower inference time than normal convolution. However, our method achieves the best performance in the pixel-oriented criteria with 215 FPS. Fu \emph{et al.} \cite{realtime} is the fastest real-time method for SR, and our method is a slightly faster than it. The experimental results demonstrate that our method is able to show the competitive speed while maintaining high performance. 

\section{Conclusion}

In this paper, we have proposed EIPNet for Face Super-Resolution. We have presented the edge block to mitigate blurring effect and LCE for global fine-tuning to improve pixel-wise accuracy. Moreover, we have provided identity loss to reconstruct SR images while preserving these personal identities. Thanks to the edge block, identity loss, and LCE, EIPNet is able to generate high quality SR images while reconstructing the high-frequency components and personal identities. The extensive experiments and ablation studies demonstrate the effectiveness of EIPNet. Therefore, EIPNet shows competitive results in pixel-wise accuracy and runtime with the state-of-the-art methods. However, EIPNet does not achieve the lowest computational complexity, and we will investigate improving it for EIPNet while increasing perceptual quality of SR images.

\section*{Acknowledgment}

This research was supported by the MSIT(Ministry of Science, ICT), Korea, under the High-Potential Individuals Global Training Program)(2019-0-01609) supervised by the IITP(Institute for Information Communications Technology Planning Evaluation); the National Research Foundation of Korea(NRF) grant funded by the Korea government (MSIT) (No. 2020R1A2C1012159); the National Natural Science Foundation of China (No. 61872280).

\bibliography{mybibfile}

\end{document}